\documentclass[runningheads]{llncs}
\usepackage{graphicx}

\usepackage{tikz}
\usepackage{comment}
\usepackage{amsmath}
\usepackage{amssymb}
\usepackage{booktabs}
\usepackage{multirow}
\usepackage{multicol}
\usepackage{algorithm}
\usepackage{algorithmic}
\usepackage{bm}
\usepackage{cite}
\usepackage{rotating}
\usepackage{makecell}

\usepackage[accsupp]{axessibility}  

\usepackage[pagebackref,breaklinks,colorlinks]{hyperref}
\usepackage{pifont}
\newcommand{\cmark}{\textcolor{red}{\ding{51}}}%
\newcommand{\xmark}{\textcolor{teal}{\ding{55}}}%

\usepackage{orcidlink} 
\usepackage{xspace}
\makeatletter
\DeclareRobustCommand\onedot{\futurelet\@let@token\@onedot}
\def\@onedot{\ifx\@let@token.\else.\null\fi\xspace}

\def\eg{\emph{e.g}\onedot} 
\def\ie{\emph{i.e}\onedot}

\def\wrt{w.r.t\onedot} 

\def\etal{\emph{et al}\onedot}
\makeatother

\usepackage[capitalize]{cleveref}
\crefname{section}{Sec.}{Secs.}
\Crefname{section}{Section}{Sections}
\Crefname{table}{Table}{Tables}
\crefname{table}{Tab.}{Tabs.}
\newcommand{\Loss}{\mathcal{L}}
\usepackage{xcolor}
\usepackage[normalem]{ulem}

\begin{document}
\pagestyle{headings}
\mainmatter
\def\ECCVSubNumber{1534}  

\title{Self-supervised Human Mesh Recovery with Cross-Representation Alignment} 
\titlerunning{Self-supervised Human Mesh Recovery with Cross-Representation Alignment}

\author{
Xuan Gong\inst{1,2}\orcidlink{0000-0001-8303-633X} \and Meng Zheng\inst{2}\orcidlink{0000-0002-6677-2017} \and
Benjamin Planche\inst{2}\orcidlink{0000-0002-6110-6437} \and Srikrishna Karanam\inst{2}\orcidlink{0000-0002-7627-7765} \and
 \\
Terrence Chen\inst{2} \and
David Doermann\inst{1}\orcidlink{0000-0003-1639-4561} \and
Ziyan Wu\inst{2}\orcidlink{0000-0002-9774-7770}}
\authorrunning{Xuan Gong et al.}
\institute{University at Buffalo, Buffalo NY, USA \\
\email{xuangong@buffalo.edu, doermann@buffalo.edu} 
\and
United Imaging Intelligence, Cambridge MA, USA \\
\email{\{first.last\}@uii-ai.com}}

\maketitle

\begin{abstract}
Fully supervised human mesh recovery methods are data-hungry and have poor generalizability due to the limited availability and diversity of 3D-annotated benchmark datasets. Recent progress in self-supervised human mesh recovery has been made using synthetic-data-driven training paradigms where the model is trained from synthetic paired 2D representation (\eg, 2D keypoints and segmentation masks) and 3D mesh. However, on synthetic dense correspondence maps (\ie, IUV) few have been explored since the domain gap between synthetic training data and real testing data is hard to address for 2D dense representation. To alleviate this domain gap on IUV, we propose cross-representation alignment utilizing the complementary information from the robust but sparse representation (2D keypoints). Specifically, the alignment errors between initial mesh estimation and both 2D representations are forwarded into regressor and dynamically corrected in the following mesh regression. This adaptive cross-representation alignment explicitly learns from the deviations and captures complementary information: robustness from sparse representation and richness from dense representation. We conduct extensive experiments on multiple standard benchmark datasets and demonstrate competitive results, helping take a step towards reducing the annotation effort needed to produce state-of-the-art models in human mesh estimation.
\end{abstract}

\section{Introduction}
\label{sec:intro}
3D human analysis from images is an important task in computer vision, with a wide range of downstream applications such as healthcare\cite{karanam2020towards} and computer animation \cite{liu2021nech}.
We consider the problem of human mesh estimation, \ie, estimating the 3D parameters of a parametric human mesh model given input data, typically RGB images. With the availability of models such as SMPL \cite{loper2015smpl}, there has been much recent progress in this area \cite{bogo2016keep, kanazawa2018end, kolotouros2019learning}. 

\begin{figure}
    \centering
    \includegraphics[width=0.7\columnwidth,trim={0 0 0 0},clip]{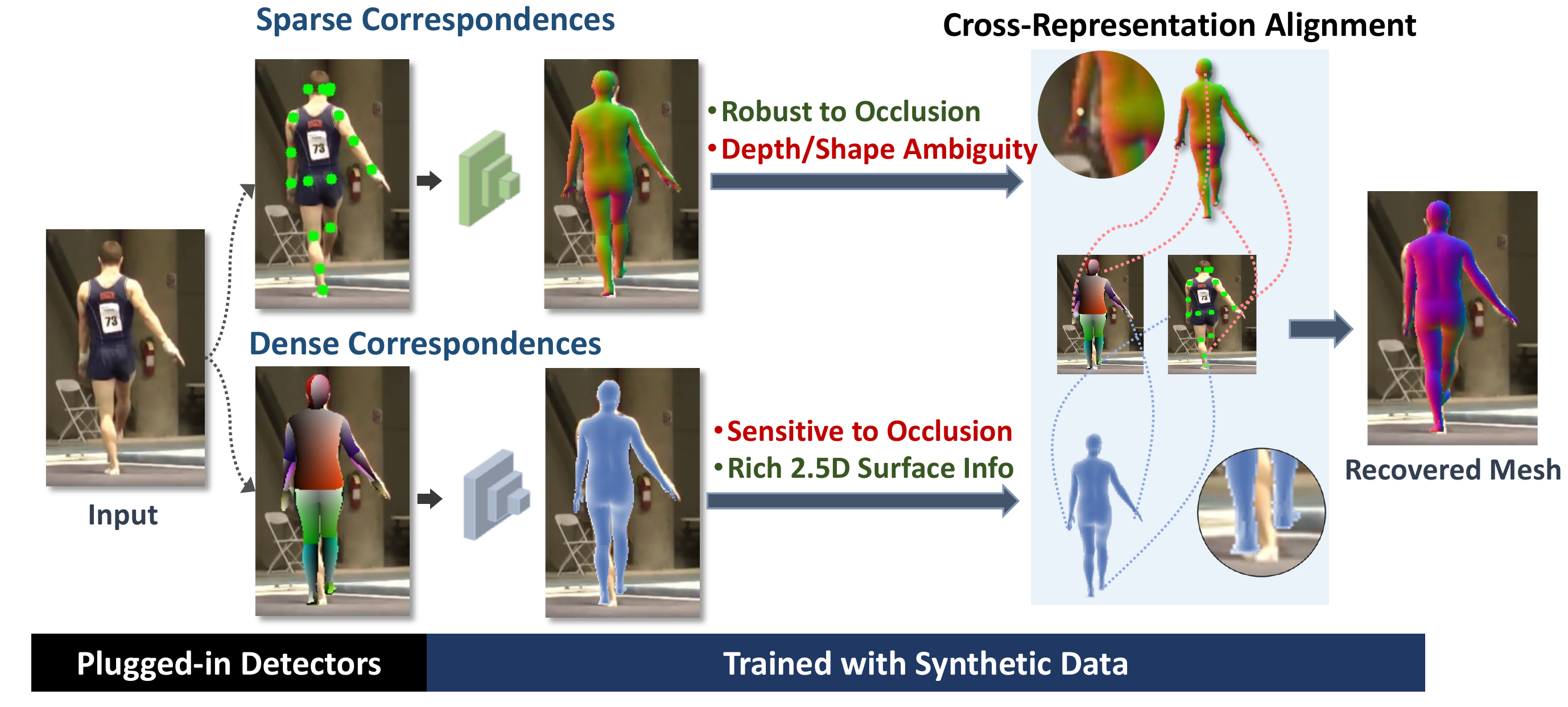}
    \caption{\textbf{Motivation:} In our synthetic-data-driven pipeline, we train a model from 2D representations to 3D mesh. During test, 2D representations are inferred from off-the-shelf detectors, where
    sparse/dense 2D representations come with complementary advantage: 2D keypoints provide a robust but sparse representation of the skeleton, dense correspondences (IUV maps) provide rich but sensitive body information. This motivates us to explore cross-representation alignment to take advantage of both to optimize recovered human mesh. }
    \label{fig1}
\end{figure}

However, obtaining good performance with these methods requires many data samples with 3D annotations. In the SMPL model, this would be the pose and shape parameters. Generating these 3D annotations is very expensive in general and prohibitive in many specific situations, such as medical settings \cite{zheng2022self}. Developing these annotations requires expensive, and custom motion capture setups and heavily customized algorithms such as MoSh \cite{loper2014mosh}, which are highly impractical in many scenarios, including the aforementioned medical one. 
This results in a situation where there are only limited datasets with 3D pose and shape annotations, further resulting in models that tend to perform well in narrow scenarios while generalizing poorly to out of distribution data \cite{liu2022simultaneously}. 

To relieve the requirement of expensive 3D labels, attempts are made to utilize more easily obtained annotations, \eg, 2D landmarks and silhouettes \cite{tan2017indirect, pavlakos2018learning, rong2019delving, Wehrbein_2021_ICCV}, ordinal depth relations \cite{pavlakos2018ordinal}, dense correspondences \cite{guler2019holopose}, or 3D skeletons \cite{Li_2021_CVPR}. To get rid of weak supervision, some take a step further by exploring temporal or multi-view images \cite{kundu2020self, wandt2021canonpose} or prior knowledge such as poses with temporal consistency \cite{yu2021towards}.

There has been some recent works  using synthetic data for human body modeling, \eg, dense correspondences estimation \cite{zhu2020simpose}, depth estimation \cite{varol2017learning}, 3D pose estimation \cite{rogez2016mocap, varol2017learning, kundu2020self, patel2021agora, song2021human}, and 3D human reconstruction \cite{zheng2019deephuman}. While these approaches show promising results, they need to render images under various synthetically designed conditions such as lighting and background. However, it is very challenging for such an approach to produce data (and hence the resulting trained model) that generalizes to real-world conditions. 
In contrast, \cite{song2020human, sengupta2020synthetic, yu2021skeleton2mesh, Sengupta_2021_CVPR, Sengupta_2021_ICCV, clever2022bodypressure} rely on various intermediate representations used for adjacent tasks such as keypoint, binary silhouettes, edges, and depth. Concretely, while insufficient data handicaps 3D human mesh estimation, tasks such as keypoint estimation have substantially more annotated data. This then leads to a situation where one can expect intermediate representations for these tasks (\eg, 2D keypoints estimation, binary silhouettes) to generalize better than the representation learned by standard mesh estimation models such as SPIN \cite{kolotouros2019learning}. At test time on real data, all one needs to do is to compute these representations with off-the-shelf detectors and then infer with the trained intermediate-representation-to-mesh regressor. 

Although the aforementioned synthesis-based methods regress the SMPL parameters directly from intermediate representations such as 2D keypoints, binary silhouettes, and depth, none of them  successfully utilize synthetic dense correspondence maps
(\ie, IUV), which can provide richer and complementary information to 2D joints/edge/silhouette. While adding IUV
to the representations may seem incremental, \cite{sengupta2020synthetic} acknowledge it is actually challenging due to the large domain gap between real IUV and synthetic IUV.

\begin{figure*}[t]
\begin{center}
  \includegraphics[width=\linewidth]{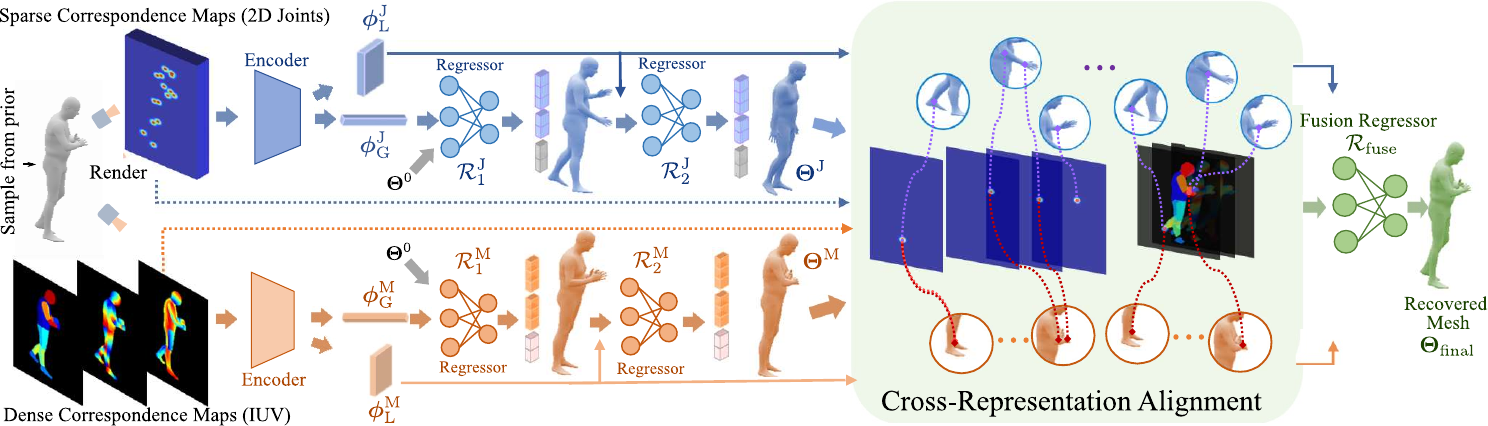}
\end{center}
   \caption{Overview of the proposed pipeline with cross-representation alignment. For training, we generate paired data between 3D mesh and intermediate representations (\ie 2D joints and IUV map).  }
\label{fig2}
\end{figure*}

We propose cross-representation alignment (CRA) to address the large domain gap while employing dense intermediate representation in synthetic training to handle all the above considerations. Our critical insight is that all these representations may not be wholly consistent but come with complementary advantages. For instance, while 2D keypoints provide a robust sparse representation of the skeleton, dense correspondences (via UV maps) can help further finetune/finesse the final output (shown in Figure \ref{fig1}). 

To this end, our proposed CRA fusion module comprises a trainable alignment scheme between the regressed mesh output and the evidential representations as part of an iterative feedback loop  (shown in Figure \ref{fig2}). 
Unlike our counterparts \cite{song2020human, sengupta2020synthetic, yu2021skeleton2mesh} which simply concatenate the features from each representation and regress SMPL parameters iteratively. We instead exploit the complementary information among different representations by generating feedback based on alignment error between the mesh estimation and each representation. The alignment feedback is then forwarded into the following regressor inferring the final SMPL estimation. By introducing the feedback mechanism here, our proposed method can effectively exploit the complementary knowledge between both representations and adapt to their different characteristics, not only during training but also after deployment with real data.
  
To summarize, our key contributions are:
\begin{itemize}
    \item We propose a novel synthetic-training pipeline successfully utilizing both sparse and dense representation by bridging the synthetic-to-real gap in dense correspondence via adaptive representation alignment.
    \item We capture complementary advantages in cross-modality with a trainable cross-representation fusion module that aligns the regressed mesh output with representation evidence as part of the iterative regression.
    \item We conduct extensive benchmarking on standard datasets and demonstrate competitive numbers with conventional evaluation metrics and protocols.
\end{itemize}

\section{Related Work}
\textbf{Single-image human 3D pose/mesh estimation.}
The emergence of statistical body models such as SCAPE \cite{anguelov2005scape} and SMPL \cite{loper2015smpl}  makes it possible to represent the human body with low-dimensional parameters. Iterative optimization-based approaches have been leveraged to fit these parametric models to 2D observations such as keypoints \cite{bogo2016keep, pavlakos2019expressive} and silhouettes \cite{lassner2017unite}. These model-fitting approaches are time-consuming, sensitive to initialization, and difficult to tune. Recent advances are dominated by learning-based methods which regress a parametric model (\eg, pose and shape parameters for SMPL \cite{loper2015smpl}) or non-parametric model (\eg, mesh vertices \cite{kolotouros2019convolutional}) under the supervision of 3D labels. Several works learn 3D body mesh from image through intermediate representations, \eg, surface keypoints \cite{tan2017indirect, Wehrbein_2021_ICCV}, silhouettes \cite{pavlakos2018learning}, body part segmentations \cite{omran2018neural}, IUV maps \cite{xu2019denserac, zhang2020learning, zeng20203d}, and 3D markers \cite{Zanfir_2021_ICCV}. Others directly learn 3D body parameters from the input image \cite{kolotouros2019learning}. Recent works have explored body kinematics \cite{georgakis2020hierarchical, xu2021monocular}, pose augmentation, and pose probabilistic distributions \cite{Kolotouros_2021_ICCV} to boost performance.
Self-attention and graph convolutional networks have also been used to learn relationships among vertices~\cite{lin2021end}, body-parts \cite{Kocabas_2021_ICCV, Zou_2021_ICCV} to handle occlusions. 

\noindent
\textbf{Weakly-supervised human 3D pose/mesh estimation.}
Several works take steps to leverage a variety of easily obtained clues, such as paired 2D landmarks and silhouettes \cite{tan2017indirect, pavlakos2018learning, rong2019delving, Wehrbein_2021_ICCV}, ordinal depth relations \cite{pavlakos2018ordinal}, DensePose \cite{guler2019holopose}, 3D skeleton \cite{Li_2021_CVPR}.  HMR \cite{kanazawa2018end} fits SMPL parameters to 2D ground-truth and utilizes adversarial learning to exploit unpaired 3D data to relieve the reliance on expensive 3D ground truth.  Kundu \etal \cite{kundu2020appearance} learn human pose and shape with 2D evidence together with appearance consensus between pairs of images of the same person. Based on GHUM \cite{xu2020ghum} as the parametric model, THUNDR \cite{Zanfir_2021_ICCV} realizes weak-supervision via intermediate 3D marker representation.

\noindent
\textbf{Self-supervised human 3D pose/mesh estimation.}
Kundu \etal \cite{kundu2020appearance, kundu2020self} utilize temporal and multi-view images as pairs and  background/foreground disentangling for self-supervision of human pose/mesh estimation. Multi-view self-supervised 3D pose estimation methods \cite{rhodin2018unsupervised, kocabas2019self, wandt2021canonpose} usually require additional knowledge \wrt the scene and camera position or multi-view images. In the absence of multi-view video sequences and other views, geometric consistency \cite{chen2019unsupervised}, kinematics knowledge \cite{kundu2020kinematic}, and temporally consistent poses  \cite{yu2021towards} have been explored for auxiliary prior self-supervision. HUND \cite{zanfir2021neural} utilizes in-the-wild images and learns the mesh with differential rendering measures between predictions and image structures. Other synthesis-based methods generate 2D keypoints, silhouettes \cite{sengupta2020synthetic, Sengupta_2021_CVPR, Sengupta_2021_ICCV}, and 3D skeleton \cite{yu2021skeleton2mesh} with existing MoCap data for training. 

\section{Method}
\subsection{Prerequisites}
\textbf{3D Human Mesh Parameterization:} We parameterize the 3D human mesh using the Skinned Multi-Person Linear (SMPL) model. SMPL\cite{loper2015smpl} is a parametric model providing independent body shape $\bm{\beta}$ and pose $\bm{\theta}$ representations with low-dimensional parameters (\ie, $\bm{\beta} \in \mathbb{R}^{10}$ and $\bm{\theta} \in \mathbb{R}^{72}$). Pose parameters include global body rotation (3-DOF) and relative 3D rotations of 23 joints (23$\times$3-DOF) in the axis-angle format. The shape parameters indicating individual heights and weights (among other parameters) are the first 10 coefficients of a PCA shape space. SMPL provides a differentiable kinematic function $\mathcal{S}$ from these pose/shape parameters to 6890 mesh vertices: $\bm{v}=\mathcal{S}(\bm{\theta}, \bm{\beta}) \in \mathbb{R}^{6890 \times 3}$. Besides, 3D joint locations for $N_\text{J}$ joints of interest are obtained as $\bm{j}^\text{3D}=\mathcal{J}\bm{v}$, where $\mathcal{J} \in \mathbb{R}^{N_\text{J} \times 6890}$ is a learned linear regression matrix.

\noindent
\textbf{Dense Human Body Representation:} We use DensePose \cite{guler2018densepose} to establish dense correspondence between the 2D image and the mesh surface behind clothes. It semantically defines 24 body parts as $\bm{I}$ to represent Head, Torso, Lower/Upper Arms, Lower/Upper Legs, Hands and Feet, where head, torso, and lower/upper limbs are partitioned into frontal-back parts to guarantee body parts are isomorphic to a plane. For UV parametrization, each body part index has a unique UV coordinate which is geometrically consistent. In this manner, with IUV representation each pixel can be projected back to vertices on the template mesh according to a predefined bijective mapping between the 3D surface space and the  IUV space. We denote the IUV map as $[\bm{I}, \bm{U}, \bm{V}] \in \mathbb{R}^{3 \times (P+1) \times H \times W}$, where $P=24$ indicating 24 foreground body parts, $H$ and $W$ are the height and width of IUV map. The index channel is one-hot indicating whether it belongs to the background or specific body part: $\bm{I} \in \{0,1\}^{(P+1) \times H \times W}$. While $\bm{U}$ and $\bm{V}$ are independent channels containing the U, V values (ranging from 0 to 1) for corresponding body part \cite{guler2018densepose}.  IUV can be further reorganized as a more compact representation $\bm{M}=[\bm{M}^\text{I}, \bm{M}^\text{U}, \bm{M}^\text{V}] \in \mathbb{R}^{3 \times H \times W}$ which is convertible with the explicit one-hot IUV version mentioned above. With $h=1,\ldots,H$ and $w=1,\ldots,W$ as pixel position, we have $M_{hw}^I \in \{0,1,\ldots,P\}$, where $0$ indicates background and non-zero value indicates body part index. As at most one out of the $P+1$ channels (background and body parts) has non-zero U/V values, the simplified $\bm{M}^\text{U}$ and $\bm{M}^\text{V}$ are represented by $M_{hw}^\text{U}=U_{M_{hw}^\text{I} hw}$, $M_{hw}^\text{V}=V_{M_{hw}^\text{I} hw}$.

\subsection{Training Data Synthesis}
\label{sec: trdatasyn}
We generate paired 2D representations and 3D meshes on-the-fly with SMPL. We utilize prior poses from the existing MoCap \cite{mocap,rogez2016mocap} datasets for diverse and realistic simulation. Body shape parameters are sampled from normal distribution $\beta_n \sim \mathcal{N}(\mu_n, \sigma_n^2) (n=1,\ldots,10)$, where the mean and variance are empirically obtained from prior statistics \cite{sengupta2020synthetic} for generalization. We employ perspective  projection with identity camera rotation $\bm{r} \in \mathbb{R}^{3 \times 3}$, dynamically sampled camera translation $\bm{t} \in \mathbb{R}^{3}$ as extrinsic parameters, and fixed focal length $\bm{f} \in \mathbb{R}^{2}$ as intrinsic parameters.

At each training step, the sampled $\bm{\theta}$ and $\bm{\beta}$  are forwarded into SMPL model to obtain mesh vertex $\bm{v}$ and 3D joints $\bm{j}^\text{3D}$. 
Then we project the 3D joints $\bm{j}^\text{3D}$ to 2D joints $\bm{j}^\text{2D}$, with sampled extrinsic and intrinsic camera parameters mentioned above: $\bm{j}^\text{2D} = \bm{f} \rm\Pi$ $(\bm{r} \bm{j}^\text{3D}+\bm{t})$,
where $\rm\Pi$ denotes perspective projection. We normalize the $\bm{j}^\text{2D}$ to be from -1 to 1, and denote normalized version as $\bm{j}^\text{2D}$ in the following for simplification.  
With these camera parameters, we render the human mesh to 2D dense IUV  based on an existing rendering method \cite{ravi2020pytorch3d}. Specifically, we take predefined unique IUV value for each vertex on the SMPL model as a template, project the vertex IUV into 2D and then obtain a continuous 2D IUV map via rasterization and shading.

The 2D joints $\bm{j}^\text{2D} \in \mathbb{R}^{N_\text{J} \times 2}$ are transformed into 2D Gaussian joint heatmaps $\bm{J} \in \mathbb{R}^{N_\text{J} \times H \times W}$ as inputs to our neural networks.  The IUV map with $\bm{M} \in \mathbb{R}^{3 \times H \times W}$ is used as the other 2D representation. Note that we normalize the I channel in $\bm{M}$ to values between $[0,1]$. For simplification, we subsequently denote the normalized version as $\bm{M}$. Finally, we have the synthesized paired data with 2D representations $\{ \bm{j}^\text{2D}, \bm{J}, \bm{M}\}$ and 3D mesh $\{ \bm{\theta}, \bm{\beta}, \bm{v}, \bm{j}^\text{3D}\}$.

\subsection{Individual Coarse-to-fine Regression}
\label{sec: individualregress}
Given the 2D representation (either $\bm{J}$ or $\bm{M}$), we first extract features with an encoder, then forward the features into the regressor, and predict the SMPL model with pose, shape, and camera parameters $\bm{\Theta} = \{ \hat{\bm{\theta}}, \hat{\bm{\beta}}, \hat{\bm{\pi}}\}$.

The encoder takes 2D representation as input and outputs features $\bm{\phi}_0 \in \mathbb{R}^{C_0 \times H_0 \times W_0}$. Before forwarding the features into the following regressor, we reduce the feature dimensions spatial-wisely and channel-wisely to maintain more global and local information. 
For global features, we use average-pooling to reduce spatial dimension and get $\bm{\phi}_\text{G} = \text{AvgPool}(\bm{\phi}_0) \in \mathbb{R}^{C_0 \times 1 \times 1}$.
For fine-grained features, we use a multi-layer perceptron (MLP) for channel reduction and retain the spatial dimension the same:
\begin{equation}
    \bm{\phi}_l = \left\{
    \begin{aligned}
    &\mathcal{P}_l(\bm{\phi}_{l-1})  &\mathrm{if~} l=1\\
    &\mathcal{P}_l(\bm{\phi}_{l-1} \oplus \bm{\phi}_0) &\mathrm{~~if~} l>1,
    \end{aligned}
    \right.
\end{equation}
where $\oplus$ denotes concatenation, $l=1,\ldots,L$ is the perception layer, $\mathcal{P}_l$ indicates the $l$-th perceptron, and $\bm{\phi}_l \in \mathbb{R}^{C_l \times H_0 \times W_0}$ with channel $C_l$ monotonically decreasing.
We denote the final output after MLP as $\bm{\phi}_\text{L}$. 

Taking the flattened feature $\bm{\phi}$ and initialized $\bm{\Theta}^0$  as input, the regressor $\mathcal{R}$ updates $\bm{\Theta}=\{ \hat{\bm{\theta}}, \hat{\bm{\beta}}, \hat{\bm{\pi}}\}$.
Note here that we use continuous 6-dimensional representation \cite{zhou2019continuity} for optimization of joint rotation in $\hat{\bm{\theta}} \in \mathbb{R}^{24\times 6}$ which can be converted to the discontinuous Euler rotation vectors. The predicted camera parameters for the standard weak-perspective projection are represented by $\hat{\bm{\pi}}=[\hat{{\pi}}_\text{s}, \hat{\bm{\pi}}_\text{t}]$, where $\hat{{\pi}}_\text{s} \in \mathbb{R}$ is the scale factor and $\hat{\bm{\pi}}_\text{t} \in \mathbb{R}^2$ indicates translation. 
Similar to the standard iterative error feedback (IEF) procedure \cite{kanazawa2018end}, we iteratively update the prediction $\bm{\Theta}$. For each representation ($\bm{J}$ and $\bm{M}$) stream, we have two regressors $\mathcal{R}_1$ and $\mathcal{R}_2$ estimating $\bm{\Theta}$ with global feature $\bm{\phi}_\text{G}$ and fine-grained feature $\bm{\phi}_\text{L}$ respectively:
\begin{equation}
\bm{\Theta}^\text{J} =  \mathcal{R}_2^\text{J} (\bm{\phi}_\text{L}^\text{J}; \mathcal{R}_1^\text{J} (\bm{\phi}_\text{G}^\text{J}; \bm{\Theta}^0)) 
\qquad \text{and} \qquad 
\bm{\Theta}^\text{M} =  \mathcal{R}_2^\text{M} (\bm{\phi}_\text{L}^\text{M}; \mathcal{R}_1^\text{M} (\bm{\phi}_\text{G}^\text{M}; \bm{\Theta}^0)) ,
\end{equation}
where $\bm{\Theta}^\text{J}$ and $\bm{\Theta}^\text{M}$ are the parameter predictions for the 2D joints representation $\bm{J}$ and IUV representation $\bm{M}$ respectively.

\subsection{Evidential Cross-Representation Alignment}
\label{sec: crossalign}
To utilize the complementary information of both representations, we design a novel fusion module $\mathcal{R}_\text{fuse}$ considering the misalignment between the prediction and the evidence from the intermediate representations (\ie, 2D joints and IUV map). One observation is that the pose parameters are represented as relative rotations and kinematic trees where minor parameter differences can result in significant misalignment on 2D projections. Another observation is that the inferred 2D joints and IUV map are likely to be noisy and inconsistent in real scenarios. During testing, we can hardly distinguish which of the available 2D representations is more reliable, so we incorporate alignment between both pieces of evidence and both predictions.

Given $\bm{\Theta} = \{ \hat{\bm{\theta}}, \hat{\bm{\beta}}, \hat{\bm{\pi}}\}$ as prediction, SMPL takes $\hat{\bm{\theta}}$ and $\hat{\bm{\beta}}$ to output 3D vertices $\hat{\bm{v}}$ and 3D joints $\hat{\bm{j}}^\text{3D}$. Then with predicted camera parameters $\hat{\bm{\pi}}$, we have the reprojected 2D joints $\hat{\bm{j}}^\text{2D} = \hat{{\pi}}_s \rm \Pi$ $(\hat{\bm{j}}^\text{3D}) + \hat{\bm{\pi}}_t$ with orthographic projection function $\rm{\Pi}$. We denote normalized version of $\hat{\bm{j}}^\text{2D}$ as $\hat{\bm{j}}^\text{2D}$ in the following for simplification.
We also render the IUV map $\widehat{\bm{M}} \in \mathbb{R}^{3 \times H_0 \times W_0}$ with $\hat{\bm{v}}$, $\hat{\bm{\pi}}$ and predefined unique IUV value for each vertex on the SMPL. Note that our projections and rendering techniques are differentiable. 

To evaluate the misalignment on 2D joints, we have 
\begin{equation}
\label{eq: j2ddiff}
    \mathcal{D}_\text{J} (\hat{\bm{j}}^\text{2D}, {\bm{j}}^\text{2D})= \hat{\bm{j}}^\text{2D} - {\bm{j}}^\text{2D},
\end{equation}
where $\mathcal{D}_\text{J}(\cdot, \cdot) \in \mathbb{R}^{N_\text{J} \times 2}$ is a discrepancy vector which can  also be seen as 2D joints pixel index offset between the prediction and the evidence. For misalignment between predicted IUV map $ \widehat{\bm{M}} = [\widehat{\bm{M}}^\text{I}, \widehat{\bm{M}}^\text{U}, \widehat{\bm{M}}^\text{V}]$ and evidential IUV map ${\bm{M}} \in \mathbb{R}^{3 \times H \times W}$, we downsize $\bm{M}$ to be with $\mathbb{R}^{3 \times H_0 \times W_0}$. For simplicity, we use ${\bm{M}}$ to represent the downsized version from now on.
The discrepancy map $\mathcal{D}_\text{M}(\cdot, \cdot) \in \mathbb{R}^{H_0 \times W_0}$ can be obtained:
\begin{equation}
\label{eq: iuvdiff}
\begin{aligned}
\mathcal{D}_\text{M} (\widehat{\bm{M}}, {\bm{M}})=  \frac{|\widehat{\bm{M}}^\text{I}-{\bm{M}}^\text{I}|}{|\widehat{\bm{M}}^\text{I}-{\bm{M}}^\text{I}|_\text{d} + \epsilon} 
&+ \sum_{p=1}^{P}{[\mathbf{1}(\widehat{\bm{M}}^\text{I}\text{=}\frac{p}{P}) \odot \widehat{\bm{M}}^\text{U} -\mathbf{1}( {\bm{M}}^\text{I}\text{=}\frac{p}{P}) \odot {\bm{M}}^\text{U} ]}  \\
&+\sum_{p=1}^{P}{[\mathbf{1}(\widehat{\bm{M}}^\text{I}\text{=}\frac{p}{P}) \odot \widehat{\bm{M}}^\text{V} -\mathbf{1}( {\bm{M}}^\text{I}\text{=}\frac{p}{P}) \odot {\bm{M}}^\text{V} ]},
\end{aligned}
\end{equation}
where the $|\cdot|$ indicates $\ell_1$ norm, $(\cdot)_\text{d}$ indicates detachment from gradients, $\epsilon = 1e^{-5}$ is to prevent the denominator to be zero; thus the first term corresponds to a differentiable version of the indicator function $\mathbf{1}(\widehat{\bm{M}}^\text{I} \text{=}  \widetilde{\bm{M}}^\text{I})$. In the second and third terms, $P=24$ indicates the 24 body parts, $\odot$ denotes element-wise multiplication, and the indicator function $\mathbf{1}$ judges whether $\widetilde{\bm{M}}^\text{I}$ or $\widehat{\bm{M}}^\text{I}$ corresponds to specific body part $p$, which is normalized here as $\frac{p}{P}$.

To simplify the notations, from this point on, we refer to $\hat{\bm{j}}^\text{2D}$ as $\hat{\bm{j}}$, we have $\{\hat{\bm{j}}^\text{J}, \widehat{\bm{M}}^\text{J}\}$ and $\{\hat{\bm{j}}^\text{M}, \widehat{\bm{M}}^\text{M}\}$ corresponding to $\bm{\Theta}^\text{J}$ and $\bm{\Theta}^\text{M}$ respectively. Then we have $\bm{D}_\text{J}^\text{J} =\mathcal{D}_\text{J}(\hat{\bm{j}}^\text{J}, \bm{j}^\text{2D})$
and $\bm{D}_\text{J}^\text{M} = \mathcal{D}_\text{J}(\hat{\bm{j}}^\text{M}, \bm{j}^\text{2D})$ as the 2D joints misalignment between the two predictions and the evidence. And $\bm{D}_\text{M}^\text{J} =\mathcal{D}_\text{M} (\widehat{\bm{M}}^\text{J}, {\bm{M}})$ and $\bm{D}_\text{M}^\text{M} = \mathcal{D}_\text{M} (\widehat{\bm{M}}^\text{M}, {\bm{M}})$ as the IUV misalignment between the two predictions and the evidences. All these misalignment representations are flattened and then taken as input of $\mathcal{R}_\text{fuse}$ along with the flattened features $\bm{\phi}_\text{L}^\text{J}$ and $\bm{\phi}_\text{L}^\text{M}$:
\begin{equation}
\bm{\Theta}^\text{final} = \mathcal{R}_\text{fuse}(\bm{D}_\text{J}^\text{J},\bm{D}_\text{M}^\text{J},\bm{\phi}_\text{L}^\text{J}, \bm{D}_\text{J}^\text{M},\bm{D}_\text{M}^\text{M}, \bm{\phi}_\text{L}^\text{M}; \bm{\Theta}^\text{J}, \bm{\Theta}^\text{M}),
\end{equation}
where $\bm{\Theta}^\text{final}$ is the final prediction initialized with both $\bm{\Theta}^\text{J}$ and  $\bm{\Theta}^\text{M}$. Note that each step of the fusion module is differentiable, \ie, maintaining the gradients so that the following loss function is able to penalize misalignment and correct the precedent prediction from $\mathcal{R}_1^\text{J}$,  $\mathcal{R}_2^\text{J}$,  $\mathcal{R}_1^\text{M}$,  $\mathcal{R}_2^\text{M}$ during training.

\subsection{Loss Function}
As described in Section \ref{sec: crossalign}, from $\bm{\Theta}^\text{final}$ we can obtain predicted vertices $\hat{\bm{v}}$, 3D joints $\hat{\bm{j}}^\text{3D}$, and project to 2D joints $\hat{\bm{j}}^\text{2D}$. We have prediction and supervision in terms of vertices, 2D joints, 3D joints and SMPL parameters respectively.
To balance among these parts, we make the loss weights learnable using homoscedastic uncertainty as in prior works \cite{kendall2018multi, sengupta2020synthetic}:
\begin{equation}
\label{eq: loss}
\begin{aligned}
&\Loss_\text{reg} (\hat{\bm{v}}, \hat{\bm{j}}^\text{2D}, \hat{\bm{j}}^\text{3D}, \hat{\bm{\theta}}, \hat{\bm{\beta}}, \bm{v}, {\bm{j}}^\text{2D}, {\bm{j}}^\text{3D}, {\bm{\theta}}, {\bm{\beta}} ) \\
= &\frac{ \Loss_2 (\hat{\bm{v}}, \bm{v})}{\sigma_\text{v}^2} +\frac{\Loss_2( \hat{\bm{j}}^\text{2D}, {\bm{j}}^\text{2D})}{\sigma_\text{j2D}^2} +\frac{\Loss_2( \hat{\bm{j}}^\text{3D}, {\bm{j}}^\text{3D})}{\sigma_\text{j3D}^2} \\
& +\frac{\Loss_2([\hat{\bm{\theta}}, \hat{\bm{\beta}}], [{\bm{\theta}}, {\bm{\beta}}])}{\sigma_\text{SMPL}^2} 
    +\text{log}(\sigma_\text{v}\sigma_\text{j2D}\sigma_\text{j3D}\sigma_\text{SMPL}),
\end{aligned}
\end{equation}
where $\Loss_2$ denotes the mean square error (MSE), and $\sigma_\text{v}$, $\sigma_\text{j2D}$, $\sigma_\text{j3D}$ and $\sigma_\text{SMPL}$ indicates weights for vertex, 2D joints, 3D joints, SMPL parameters which are adaptively adjusted during training.

\textbf{Auxiliary Refinement.}
Our framework can naturally refine the network with available in-the-wild images. Given an image, we use an existing off-the-shelf detector to obtain IUV map $\bm{M}$ and 2D joints $ \bm{j}^\text{2D}$. The IUV map is downsampled and the 2D joints are processed to Gaussian heatmaps $\bm{J}$.  We take $\{\bm{M},\bm{J} \}$ as input, forward through our network, and output the final prediction $\bm{\Theta}$. As described in Section \ref{sec: crossalign}, we obtain the reprojected $\hat{\bm{j}}^\text{2D}$ and rendered $\widehat{\bm{M}}$ in a differentiable manner. Given $\mathcal{D}_\text{M}$ defined in Equation \ref{eq: iuvdiff}, the refinement loss function is thus computed as:
\begin{equation}
\label{eq: lossrefine}
    \Loss_\text{refine} (\hat{\bm{j}}^\text{2D}, \widehat{\bm{M}}, {\bm{j}}^\text{2D}, {\bm{M}}) = \Loss_\text{2}(\hat{\bm{j}}^\text{2D}, {\bm{j}}^\text{2D})+ \mathcal{D}_\text{M}(\widehat{\bm{M}}, {\bm{M}}),
\end{equation}

\section{Experiments}
\subsection{Datasets}
\textbf{Training data.}
To generate synthetic training data, we sample SMPL pose parameters from the training sets of UP-3D \cite{lassner2017unite}, 3DPW \cite{von2018recovering}, and the five training subjects of Human3.6M \cite{ionescu2013human3} (S1, S5, S6, S7, S8). 
The sampling of shape parameters follows the procedure of prior work \cite{sengupta2020synthetic}. 

\noindent
\textbf{Evaluation data.}
We report evaluation results on both indoor and outdoor datasets, including 3DPW \cite{von2018recovering}, MPI-INF-3DHP \cite{mehta2017monocular}, and Human3.6M \cite{ionescu2013human3} (Protocols 1 and 2 \cite{kanazawa2018end} with subjects S9, S11). For 3DPW,
we report the mean per joint position error (MPJPE), mean per joint position error after rigid alignment with Procrustes analysis (PMPJPE), and after-scale correction \cite{sengupta2020synthetic} for pose estimation, and per-vertex error (PVE) for shape estimation.
For MPI-INF-3DHP, we report metrics after rigid alignment, including PMPJPE, percentage of correct keypoints (PCK) thresholded at 150mm, and the area under the curve (AUC) over a range of PCK thresholds \cite{mehta2017monocular}. For Human3.6M, we report MPJPE and PMPJPE on protocols 1 and 2 using the H3.6M joints definition.  

\subsection{Implementation Details}
\textbf{Synthetic data preprocessing and augmentation:} We generate paired data on-the-fly with details described in Section \ref{sec: trdatasyn}. We follow the hyperparameters in \cite{sengupta2020synthetic} for SMPL shape and camera translation sampling. We use $N_\text{J}=17$ COCO joints  to extract 3D joints from the SMPL model and then project to 2D joints representation.
The vertices $\bm{v}$ are randomly perturbed within $[-10\text{mm}, 10\text{mm}]$ for augmentation. From perturbed vertices and sampled camera parameters, we render 2D IUV map $\bm{M}$ based on Pytorch3D \cite{ravi2020pytorch3d}. We detect the foreground body area on 2D IUV and crop around the foreground area with a scale of $1.2$ around the bounding box, which is unified for consistency between training and testing. We crop both IUV $\bm{M}$ and joints heatmaps $\bm{J}$ and then resize to the target size with $H=256$, $W=256$. To simulate noise and discrepancy between 2D joints and IUV prediction, we do a series of probabilistic augmentations, including randomly masking one of the six body parts  (same as PartDrop in \cite{zhang2020learning}),  randomly masking one of the six body parts (head, torso, left/right arm,  left/right leg) on IUV map, randomly occluding the IUV map with a dynamically-sized rectangle, and randomly perturbing the 2D joints position. 

\noindent
\textbf{Architecture:}
We use ResNet-18 \cite{He_2016_CVPR} as encoder and the size of the output $\bm{\phi}_0$ is $C_0=512$, $H_0=8$, $W_0=8$. Through average pooling we get $\bm{\phi}_\text{G}$ with size  $512 \times 1 \times 1$. Each perceptron $\mathcal{P}_l$ in the MLP consists of Conv1D and ReLU operations with $L=3$ layers in total. The MLP reduce the feature channels to $[C_1, C_2, C_3]=[256,64,8]$ progressively, and produces the feature vector $\bm{\phi}_\text{L}$ with size $8 \times 8 \times 8$. Each regression network for $\{\mathcal{R}_1^\text{J}, \mathcal{R}_2^\text{J}, \mathcal{R}_1^\text{M}, \mathcal{R}_2^\text{M}\}$ consists of two fully-connected layers with 512 neurons each, followed by an output layer with 157 neurons ($\bm{\Theta}=\{ \hat{\bm{\theta}}, \hat{\bm{\beta}}, \hat{\bm{\pi}}\} \in \mathbb{R}^{24 \times 6 + 10 +3}$ as explained in Section \ref{sec: individualregress}). Taking the input vector with dimension $2\times(C_L \times H_0 \times W_0 +3\times H_0\times W_0+2\times N_\text{J})=1540$, the regression network for $\mathcal{R}_\text{fuse}$ consists of two fully-connected layers with 1,540 neurons each, followed by an output layer with 157 neurons.

\begin{table}[t]
\centering
\resizebox{\textwidth}{!}
{
\begin{tabular}{c|c|ccc|cc|cc}
\toprule
\multirow{2}{*}{Method}  &{2D} &\multicolumn{3}{c|}{Auxiliary requirements} &\multicolumn{2}{c|}{Protocol \# 1}  &\multicolumn{2}{c}{Protocol \# 2}  \\ 
&Superv. & \makecell{image\\pairs} &\makecell{multi-view\\imagery} & \makecell{temporal\\prior} &MPJPE$\downarrow$ &PMPJPE$\downarrow$ &MPJPE$\downarrow$ &PMPJPE$\downarrow$ \\ \midrule
$^*$HMR (unpaired)\cite{kanazawa2018end} &\cmark &\xmark &\xmark &\xmark &106.84 &67.45 & &66.5 \\
$^*$SPIN (unpaired)\cite{kolotouros2019learning} &\cmark &\xmark &\xmark &\xmark &- &- &- &62.0 \\
$^*$Kundu \etal \cite{kundu2020appearance} &\cmark &\cmark &\xmark  &\xmark &\textbf{86.4} & & &\textbf{58.2}\\
$^*$THUNDER \cite{Zanfir_2021_ICCV} &\cmark &\xmark &\xmark &\xmark &87.0 &62.2 &83.4 &59.7\\ \midrule
Kundu \etal \cite{kundu2020kinematic} &\xmark &\cmark  &\xmark &\xmark &- &-  &- &89.4\\
Kundu \etal \cite{kundu2020self} &\xmark &\cmark  &\cmark &\xmark  &- &- &- &85.8\\
$^*$Kundu \etal \cite{kundu2020appearance} &\xmark &\cmark  &\cmark &\xmark  &102.1 &- &- &74.1 \\
CanonPose \cite{wandt2021canonpose} &\xmark &\xmark &\cmark &\xmark  & \textbf{81.9} &- &- &53 \\ 
Yu \etal \cite{yu2021towards}  &\xmark &\xmark  &\xmark &\cmark &- &- &92.4 &\textbf{52.3} \\
 \midrule
$^*$Song \etal \cite{song2020human} &\xmark &\xmark &\xmark &\xmark &- &- &- &56.4 \\
$^*$STRAP \cite{sengupta2020synthetic} &\xmark &\xmark &\xmark &\xmark &87.0 &59.3 &83.1 &55.4 \\
$^*$HUND \cite{zanfir2021neural} &\xmark &\xmark &\xmark  &\xmark &91.8
&66.0 &- &- \\
$^*$Skeleton2Mesh \cite{yu2021skeleton2mesh} &\xmark &\xmark &\xmark &\xmark   &87.1 &\textbf{55.4}  &- &-  \\
{$^*$Ours (\emph{synthesis only})}  &\xmark &\xmark &\xmark &\xmark &87.1 &58.2 &81.3 &54.8 \\
{$^*$Ours (\emph{w/ refinement}) }  &\xmark &\xmark &\xmark &\xmark &\textbf{84.3} &57.8 &\textbf{81.0} & \textbf{53.9}\\
\bottomrule
\end{tabular}}
\caption{Comparison of our method with weakly supervised and self-supervised SOTA in terms of MPJPE and PMPJPE (both in mm) on the H3.6M Protocol \#1 and Protocol \#2 test sets. $^*$ indicates methods that can estimate more than 3D pose.   
}
\label{tab:h36m}
\end{table}

\noindent
\textbf{Training:}
With the final prediction $\bm{\Theta}_\text{final}$, we use Equation \ref{eq: loss} as a loss function to train the whole network in an end-to-end fashion. We use Adam \cite{kingma2014adam} optimizer to train for 30 epochs with a learning rate of $1e^{-4}$ and a batch size of 128. 
On the image, we predict 2D joints and IUV maps using the off-the-shelf Keypoint-RCNN \cite{he2017mask} and DensePose \cite{guler2018densepose} models.  
For auxiliary refinement, we use RGB images from the corresponding training set when testing on the Human3.6M, 3DPW, and MPI-INF-3DHP. 
We use Adam to train for ten epochs with a learning rate of $1e^{-6}$ and a batch size of 128 for auxiliary refinement.

\noindent
\textbf{Testing:} We infer 2D joints on the testing images with the pretrained Keypoint-RCNN \cite{he2017mask} with ResNet-50 backbone. We obtain the IUV prediction with pretrained DensePose-RCNN \cite{he2017mask} with ResNet-101 backbone. Since 3DPW test images may have multiple persons, we use the same protocol as \cite{kolotouros2019learning} to get the bounding box for the target person by using the scale and center information and get the 2D representations with maximum IOU with the target bounding box. We crop both the IUV maps and 2D joints heatmaps with a scale of 1.2 before forwarding them to the network for 3D mesh inference. 

\begin{table}[t]
\centering
\scriptsize
{
\begin{tabular}{cccccc}
\toprule
\multicolumn{2}{c}{Method} &PVE$\downarrow$ &MPJPE$\downarrow$ &MPJPE-SC$\downarrow$ &PMPJPE$\downarrow$  \\ \midrule
{\multirow{3}{*}{\shortstack[c]{Full Superv.}}}
&HMR \cite{kanazawa2018end} &139.3  &116.5 &- &72.6 \\
&VIBE \cite{kocabas2020vibe} &113.4 &113.4 &- &56.5 \\
&PyMAF\cite{pymaf2021} &110.1 &92.8 &- &58.9 \\
\midrule
{\multirow{3}{*}{\shortstack[c]{Weak Superv.}}} &HMR (unpaired) \cite{kanazawa2018end} &- &- &126.3 &92.0  \\
&Kundu \etal \cite{kundu2020appearance} &- &153.4 &- &89.8\\ 
&THUNDER \cite{Zanfir_2021_ICCV} &- &87.8 &- &59.9\\
\midrule
{\multirow{8}{*}{\shortstack[c]{Self Superv.}}} &Kundu \etal \cite{kundu2020appearance} &- &187.1 &- &102.7 \\
&STRAP \cite{sengupta2020synthetic} &131.4  &118.3 &99.0 &66.8\\
&HUND \cite{zanfir2021neural} &- &\underline{90.4}
&- &63.5
\\
&STRAP V2 \cite{Sengupta_2021_CVPR} &- &- &90.9 &61.0\\
&STRAP V3 \cite{Sengupta_2021_ICCV} &- &- &84.7 &59.2\\
&Song \etal \cite{song2020human} &- &- &- &\textbf{55.9} \\
&{Ours (\emph{synthesis only})} &117.4 &91.1 &\underline{80.8} &\underline{56.3}\\
&{Ours (\emph{w/ refinement})}  &115.3 &\textbf{89.1} &\textbf{79.0} &\textbf{55.9}\\
\bottomrule
\end{tabular}
}
\caption{A comparison with fully/weakly/self-supervised SOTA methods in terms of PVE, MPJPE, MPJPE-SC, and PMPJPE (all in mm) on the 3DPW test dataset. 
}
\label{tab:3dpw}
\end{table}

\subsection{Quantitative Results}
\noindent
\textbf{Human3.6M:} 
We evaluate our method on the Human3.6M \cite{ionescu2013human3} test dataset (both Protocol \#1 and Protocol \#2) and compare our method with SOTA weakly supervised methods and self-supervised methods in Table \ref{tab:h36m}. Note that the weakly supervised methods utilized paired images and 2D ground-truth such as 2D joints for supervision during training. And some self-supervised methods use auxiliary clues such as image pairs in video sequences, multi-view images, or prior knowledge of human keypoint positions on temporal sequences. Without reliance on either of these prerequisites, our method shows very competitive results compared with the prior arts with auxiliary refinement.  Among the methods not requiring auxiliary clues, \eg temporal or multi-view imagery, we achieve the best results in 3D pose estimation metrics-MPJPE Protocol \#1,  MPJPE, and PMPJPE on Protocol \#2 of the Human3.6M test set.

\noindent
\textbf{3DPW:}
On the test set of 3DPW \cite{von2018recovering}, we calculate PVE as shape evaluation metric and MPJPE, PMPJPE, MPJPE-SC \cite{sengupta2020synthetic} as pose evaluation metrics. From the comparisons in Table \ref{tab:3dpw}, we note that our method outperforms the prior arts, including those trained with 3D ground-truth (\ie, full supervision) and 2D ground-truth (\ie, weak supervision), on all metrics for pose evaluation. Although we do not rely on any annotated data, our method achieves results on shape estimation comparable to the prior arts trained with 3D annotation. 

\begin{table}[t]
\centering
\scriptsize
{
\begin{tabular}{c|c|ccc}
\toprule
Method  &Images Used &PCK$\uparrow$ &AUC$\uparrow$ &PMPJPE$\downarrow$\\ \midrule
$^*$HMR (unpaired)\cite{kanazawa2018end}  &H36M+3DHP  &77.1 &40.7 &113.2 \\
Kundu \etal \cite{kundu2020kinematic}  &H36M+3DHP &79.2 &43.4 &99.2 \\
Kundu \etal \cite{kundu2020self}   &H36M+YTube  &\textbf{83.2} &\textbf{58.7} &97.6 \\
CanonPose \cite{wandt2021canonpose}  &H36M+YTube  &77.0 &- &\textbf{70.3} \\ %
\midrule
Yu \etal \cite{yu2021towards} &3DHP &86.2 &51.7 &- \\
$^*$Skeleton2Mesh \cite{yu2021skeleton2mesh}  &3DHP &87.0 &50.8 &87.4 \\ 
$^*$SPIN (unpaired)\cite{kolotouros2019learning}   &3DHP &87.0 &48.5 &80.4 \\
{$^*$Ours (\emph{synthesis only})}   &\textbf{None} &89.4 &{54.0}  &80.2\\
{$^*$Ours (\emph{w/ refinement })}  &3DHP  &\textbf{89.7} &\textbf{55.0}  &\textbf{79.1} \\
\bottomrule
\end{tabular}}
\caption{Comparison with SOTA methods in terms of PCK, AUC, and PMPJPE (mm) after rigid alignment on the MPI-INF-3DHP test dataset. $^*$ indicates methods that can estimate more than 3D pose. Methods in the top half require training images paired with 2D ground-truth. Methods in the bottom half do not. 
}
\label{tab:3dhp}
\end{table}

\noindent
\textbf{MPI-INF-3DHP:} 
On the test set of MPI-INF-3DHP \cite{mehta2017monocular}, we consider the usual metrics PCK, AUC, and PMPJPE after rigid alignment, to evaluate the 3D pose estimation. As shown in Table \ref{tab:3dhp}, other methods heavily rely on the related human image dataset for training, and some have additional requirements on multi-view images (\ie, Human3.6M) and continuous images in temporal sequence (\ie, YouTube videos). In contrast,  our method has no such requirements and yet achieves better results on PCK than the prior arts (including weakly supervised methods). With access to the images, we can refine the network with a 0.9 mm improvement in PMPJPE.  Compared with the methods relying on both temporal and multi-view images \cite{kundu2020self, wandt2021canonpose}, our method achieves state-of-the-art PCK and very competitive AUC and PMPJPE  without any requirements of images. Notably, we do not use any prior information of MPI-INF-3DHP during synthetic training but still achieve very competitive performance on MPI-INF-3DHP with model only trained with synthetic data. This demonstrates the superiority of our method's generalization ability to unseen in-the-wild data.

\begin{table}[t]
\centering
\scriptsize
{
\begin{tabular}{c|c|c|cc}
\toprule
{Representation} &Regressor &Fusion &PVE$\downarrow$ &PMPJPE$\downarrow$ \\\midrule
$^{1}$J2D &$\mathcal{R}_1$ $\mathcal{R}_1$ $\mathcal{R}_1$ &- &181.3 &75.2\\
$^{2}$IUV &$\mathcal{R}_1$ $\mathcal{R}_1$ $\mathcal{R}_1$ &- &167.2 &83.1\\
$^{3}$J2D \& IUV &$\mathcal{R}_1$ $\mathcal{R}_1$  $\mathcal{R}_1$ &input $\oplus$ &121.3 &60.1 \\\midrule
$^{4}$J2D \& IUV &$\{\mathcal{R}_1$ $\mathcal{R}_1\}^{\times 2}$  &$\mathcal{R}_\text{fuse}$ &120.8 &61.0\\
$^{5}$J2D \& IUV &$\{\mathcal{R}_1$ $\mathcal{R}_2\}^{\times 2}$  &$\mathcal{R}_\text{fuse}$ &117.7 &59.6\\
$^{6}$J2D \& IUV &$\{\mathcal{R}_1$ $\mathcal{R}_2\}^{\times 2}$  &$^{ \lhd}\mathcal{R}_\text{fuse}$ &118.6 &58.2\\
$^{7}$J2D \& IUV &$\{\mathcal{R}_1$ $\mathcal{R}_2\}^{\times 2}$  &$^{\rhd \lhd}\mathcal{R}_\text{fuse}$ &\textbf{117.4} &\textbf{56.3}\\
\bottomrule
\end{tabular}
}
\caption{Ablations of one/two representations, concatenation fusion, two-stream fusion with regressor $\mathcal{R}_3$, and the evidential representation alignment on the 3DPW test dataset in terms of PVE and PMPJPE (mm). Here $^{\lhd}\mathcal{R}$, $^{\rhd\lhd}\mathcal{R}$ denotes the regressor taking misalignment of its preceding regressor prediction in terms of $^{\lhd}$the other/$^{\rhd\lhd}$both representation(s) as additional input. Note: no refinement applied for comparison. 
}
\label{tab:ablation}
\end{table}

\begin{table*}
\centering
\scriptsize
{
\begin{tabular}{ccccccccccc}
\toprule
&Body part occlusion prob. &0.1 &0.2 &0.3 &0.4 &0.5&0.6&0.7 &0.8 &0.9\\ \hline
\multirow{3}{*}{PVE$\downarrow$  } & IUV  &163.4 &166.1 &169.0 &171.7 &174.6 &177.0 &180.1 &182.7 &185.5\\
& IUV + J2D (wo/ CRA)  &138.8 &143.6 &145.8 &148.3 &150.7 &153.0 &155.7 &157.9 &160.3\\
& IUV + J2D (w/ CRA)  &\textbf{118.1} &\textbf{118.7} &\textbf{119.4} &\textbf{120.1} &\textbf{120.7} &\textbf{121.4} &\textbf{122.0} &\textbf{122.7} &\textbf{123.4}\\\hline 
\multirow{3}{*}{PMPJPE$\downarrow$ } & IUV  &92.9 &94.9 &97.0 &99.0 &101.1 &102.9 &105.1 &106.9 &109.1\\
& IUV + J2D (wo/ CRA)  &61.8 &62.6 &64.3 &66.2 &68.0 &69.8 &71.7 &73.3 &75.1\\
& IUV + J2D (w/ CRA)  &\textbf{56.8} &\textbf{57.3} &\textbf{57.8} &\textbf{58.2} &\textbf{58.7} &\textbf{59.2} &\textbf{59.6} &\textbf{60.1} &\textbf{60.6}\\
\midrule \midrule 
&Remove 2D joints prob. &0.1 &0.2 &0.3 &0.4 &0.5 &0.6 &0.7 &0.8 &0.9\\ \hline
\multirow{3}{*}{PVE$\downarrow$} &J2D &185.0 &193.9 &204.2 &237.8 &273.7 &306.8 &339.6 &370.9 &402.1 \\
& J2D + IUV (wo/ CRA) &150.5 &164.8 &181.9 &201.9 &222.6 &246.6 &270.1 &294.0 &318.7\\
& J2D + IUV (w/ CRA) &\textbf{127.4} &\textbf{139.8} &\textbf{153.9} &\textbf{170.2} &\textbf{188.9} &\textbf{210.1} &\textbf{232.9} &\textbf{258.0} &\textbf{284.2}\\\hline
\multirow{3}{*}{PMPJPE$\downarrow$} & J2D &88.5 &98.8 &122.5 &145.8 &169.4 &189.7 &208.5 &224.6 &239.0\\
& J2D + IUV (wo/ CRA)  &68.0 &78.4 &90.0 &102.8 &114.9 &127.5 &138.7 &148.3 &156.3\\
& J2D + IUV (w/ CRA) &\textbf{63.9} &\textbf{72.7} &\textbf{82.5} &\textbf{92.8} &\textbf{104.1} &\textbf{115.7} &\textbf{128.1} &\textbf{140.5} &\textbf{153.6}\\
\bottomrule
\end{tabular}
}
\caption{Comparisons of PVE and PMPJPE (both in mm) when adding noise on IUV/2D joints representations of 3DPW test images. We study the performances when using one/two representations and using two representations with and without CRA.}
\label{tab:testaug}
\end{table*}
\noindent
\textbf{Ablations:}
In Table \ref{tab:ablation}, we study the efficacy of our cross-representation alignment, where $\oplus$ denotes concatenate two representations as input of the encoder for fusion. From line 1 to line 3, we note that using the complementary information of 2D joints and IUV is better than using only one. The bottom half shows the results under our two-stream fusion pipeline, demonstrating the efficacy of our alignment module. 
The comparison between line 4 and line 5 shows that separate regressors taking features with different scales achieve better results than iterative regression with $\mathcal{R}_1$ only taking features with size $C_0 \times 1 \times 1$. And the incorporation of our evidential representation alignment scheme (discrepancy vector/map (Equation \ref{eq: j2ddiff}/\ref{eq: iuvdiff}) between the preceding regressor's prediction and the evidence as an additional input of the regressor) achieves further improvement (line 6 over line 7).  We can see that utilizing discrepancy on both representations before $\mathcal{R}_\text{fuse}$ achieves the best result.

To study the efficiency of our proposed cross-representation alignment, we further simulate the extremely challenging conditions by adding noise on the inferred 2D joints and IUV representations. On the IUV map, we simulate the occlusion cases by masking out one of the six coarse body parts (head, torso, left/right arm,  left/right leg) with increasing probability. For 2D joints, we remove the key joints(\ie, left and right elbow, wrist, knee, ankle) with increasing probability.
From the comparisons in Table \ref{tab:testaug}, we can see that the combination of 2D joints and IUV can outperform IUV only on both shape and pose evaluations.  Notably,  our proposed cross-representation alignment (w/ CRA) outperforms the baseline (wo/ CRA) by a large margin, especially for the cases with severe noise.

\begin{figure}[t]
    \centering
    \includegraphics[width=1\columnwidth]{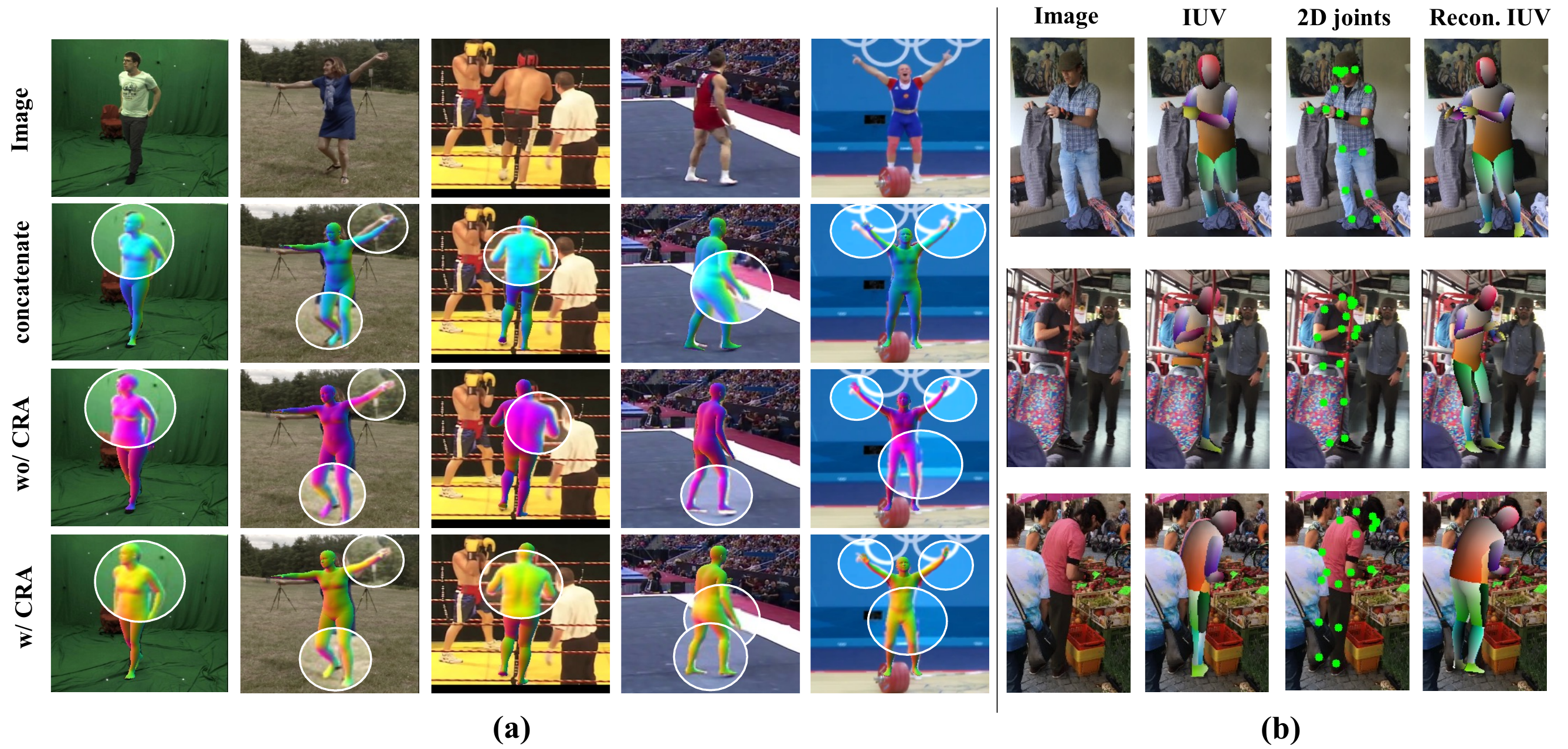}
    \caption{(a) Comparison of qualitative results on human mesh estimation: taking 2D joints and IUV as input and processing with concatenation, two-stream fusion with and without GRU.
    (b) Visualization of IUV, 2D joints and reconstructed IUV from 2D joints on 3DPW test set. Note IUV (col 1) is visualized in HSV color space, which is predicted from pretrained Densepose-RCNN. 2D joints (col 2) are predicted from Keypoint-RCNN. IUV reconstructed from the 2D joints by the decoder (col 3) is trained together with CRA.}
    \label{figvismesh}
\end{figure}

\subsection{Qualitative Results}
Qualitative examples are given in Figure \ref{figvismesh}(a). We compare our proposed CRA (row 4) with typical concatentation taking 2D joints and IUV as input (row 2), and the baseline of CRA with no alignment applied (row 3). From the highlighted part we can see that our method with alignment module achieves much better shape estimation as well as pose estimation especially on joints such as wrist and knees. Notably the visualization is on images selected from SSP-3D \cite{sengupta2020synthetic} and MPI-INF-3DHP test set of which we do not utilize any prior knowledge. The results demonstrate the robustness and generalization ability of our proposed method to unseen in-the-wild data. We observe that for a small number of cases it could be difficult for CRA to recover from errors existing in all input immediate representations (\eg, no detection on the lower body in both sparse and dense correspondences).

\textbf{Auxiliary reconstruction:} Our two-stream pipeline enables utilization of the encoded features $\bm{\phi}$ of one representation to reconstruct another representation (\eg from 2D joints to IUV map) at the same time while recovering the human mesh. We use a symmetric version of encoder as the decoder for each representation and employ the same loss function as \cite{pymaf2021} for IUV reconstruction with the synthetic IUV as supervision during training. From Figure \ref{figvismesh}(b), we note that the IUV prediction from off-the-shelf detector (trained with annotation) is occasionally sensitive to occlusion. While our recovered IUV trained with synthetic data can generalize to occlusion and more robust to ambiguous area in RGB image. 

\section{Conclusion}
We propose a novel human mesh recovery framework relying only on synthetically generated intermediate representations based on pose priors. We design a Cross-Representation Alignment module to exploit complementary features from these intermediate modalities by enforcing consistency between predicted mesh parameters and input representations. Experimental results on popular benchmark datasets demonstrate the efficacy and generalizability of this framework.



\clearpage
%
%
\bibliographystyle{splncs04}
\bibliography{egbib}
\end{document}